# Dynamic Analysis of Nonlinear Civil Engineering Structures using Artificial Neural Network with Adaptive Training

X. Pan[(1)], Z. Wen[(2)], T.Y. Yang[(3)]

[(1)] Ph.D. student, Department of Civil Engineering, University of British Columbia, p.xiao1994@outlook.com
[(2)] Undergraduate, School of Mechanical Engineering, Zhejiang University, zhizhaowen@zju.edu.cn
[(3)] Professor, Department of Civil Engineering, University of British Columbia, yang@civil.ubc.ca

*Abstract*
Dynamic analysis of structures subjected to earthquake excitation is a time-consuming process, particularly in the case of extremely small time step required, or in the presence of high geometric and material nonlinearity. Performing parametric studies in such cases is even more tedious. The advancement of computer graphics hardware in recent years enables efficient training of artificial neural networks that are well-known to be capable of learning highly nonlinear mappings. Traditional artificial neural networks are usually built on fixed architecture which consists of a fixed number of nodes and hidden layers, and consequently is more likely to get stuck in local minima during training. In this study, artificial neural networks are developed with adaptive training algorithms, which enables automatic nodes generation and layers addition. The hyperbolic tangent function is selected as the activation function. Stochastic Gradient Descent and Back Propagation algorithms are adopted to train the networks. The neural networks initiate with a small number of hidden layers and nodes. During training, the performance of the network is continuously tracked, and new nodes or layers are added to the hidden layers if the neural network reaches its capacity. At the end of the training process, the network with appropriate architecture is automatically formed. The performance of the networks has been validated for inelastic shear frames, as well as rocking structures, of which both are first built in finite element program for dynamic analysis to generate training data. Energy dissipation is considered during rocking motion using a Hilber–Hughes–Taylor numerically dissipative time step integration scheme. Results have shown the developed networks can successfully predict the time-history response of the shear frame systems subjected to real ground motion records. The response of rocking structures can be mostly well predicted although there exist small discrepancies at some instants. Furthermore, the efficiency of the proposed neural networks is also examined, which shows the computational time can be reduced by 43% than FE models, when the Graphical Processing Units acceleration is enabled. This indicates the trained networks can be utilized to generate rocking spectrums of structures more efficiently which demands a large number of time-history analyses.

# 1. Introduction

Nonlinear time-history analysis has gained its popularity in recent years to quantify the structural response more accurately. In particular, the displacement response of structures is closely related to the damage state of the structural components. The time-history response is usually computed by numerical time-stepping methods with finite element modelling. As modern computer graphics cards are specially optimized for parallel computing, artificial neural networks (ANNs) are considered as an efficient and robust computational technique for solving complex problems. The advancement of ANNs can be generally divided into 3 periods. The first wave started between 1940s–1960s, where McCulloch and Pitts [1] proposed the theories of biological learning and the first artificial neural network such as Perceptron was implemented by Rosenblatt [2]. The second phase happened from 1980 to 1995, where the back-propagation technique was developed by Rumelhart, Hinton, & Williams [3] to train a neural network with limited hidden layers. In the 1990s, the artificial neural network has evolved into deep neural networks (DNNs) which have more hidden layers and were successfully trained through the back-propagation algorithm [4]. The third phase of ANNs (also named deep learning) initiated in 2006 when Hinton, Osindero, and Teh [5] implemented a greedy layer-wise pretraining strategy in the training of deep belief network. With the fast-growing and optimization of the deep learning training algorithms and the enhanced computational power, training and testing (or deployment) of ANNs become highly efficient with the accelerated parallel computation by the graphics processing units (GPUs). ANNs have been demonstrated to provide enhanced performance compared to conventional technologies in multiple areas such as robust pattern detection, signal filtering, data segmentation, adaptive control, optimization, scheduling and complex nonlinear mapping [6]. Applications of ANNs have been widely attempted in structural engineering such as structural system identification [7], model updating of structures [8], dynamic characteristic prediction [9, 10], estimation of soil-structure interaction [11, 12]. Traditionally, ANNs are constructed using a fixed architecture with a constant number of nodes and hidden layers, and therefore are more likely to get stuck in local minima during training. In the present study, an ANN has been developed with adaptive training algorithms that can automatically generate nodes and hidden layers once the saturation is reached within the existing architecture. This can alleviate the dying neural network problem. On the other hand, as the network gradually grows itself, it avoids the use of a large and complex network, thus improving computational efficiency and allieviating overfitting. The hyperbolic tangent function is selected as the activation function. Stochastic Gradient Descent (SGD) and Back Propagation (BP) algorithms are employed to train the proposed ANNs. During training, the performance of the network is continuously monitored. At the end of the training process, the network with appropriate architecture is automatically established. The process is repeated over all the training ground motions and tested by the new ground motions. Both of the training and testing results are compared with the structural response obtained by the finite element models constructed in OpenSees [13].

# 2. Finite element models

In this section, two case studies were conducted in OpenSees. In order to examine the performance of the proposed neural networks (in Section 3) in capturing the nonlinear behaviour of the structures, the first FE model involves a nonlinear shear frame which incorporates the material nonlinearity, while the second model introduces geometric nonlinearity through the construction of a solitary rocking body. These FE models will be utilized to generate the training data and testing data for the proposed neural network model.

In the present study, the set of records is collected from the Pacific Earthquake Engineering Research Center (PEER) database. In the case of the shear frame model, in order to excite the proposed model into the highly nonlinear state, each selected ground motion is linearly scaled such that the spectral acceleration at the fundamental period of the frame is 3.0g where g is the ground acceleration. Table 1 summarizes the catalogue of earthquakes utilized in the time-history analyses of the frame. Similarly, six ground motions are utilized in the response estimation of the rocking model as shown in Table 2. To ensure reasonably large rocking amplitude and avoid the overturning scenario during rocking, the ground motions associated with the rocking model are scaled such that the peak ground acceleration (PGA) is 1.5g.

Table 1 – Ground motion selection and scaling for the shear frame model

| Record Sequence Number | Earthquake Name | Year | Magnitude | Scale Factor | Record |
|---|---|---|---|---|---|
| 15 | "Kern County" | 1952 | 7.36 | 12.22 | RSN15_KERN_TAF021 |
| 143 | "Tabas_ Iran" | 1978 | 7.35 | 2.29 | RSN143_TABAS_TAB-L1 |
| 164 | "Imperial Valley-06" | 1979 | 6.53 | 6.55 | RSN164_IMPVALL.H_H-CPE147 |
| 292 | "Irpinia_ Italy-01" | 1980 | 6.9 | 4.34 | RSN292_ITALY_A-STU000 |
| 313 | "Corinth_ Greece" | 1981 | 6.6 | 3.91 | RSN313_CORINTH_COR--L |
| 570 | "Taiwan SMART1(45)" | 1986 | 7.3 | 4.52 | RSN570_SMART1.45_45C00EW |
| 582 | "Taiwan SMART1(45)" | 1986 | 7.3 | 8.82 | RSN582_SMART1.45_45O08EW |
| 587 | "New Zealand-02" | 1987 | 6.6 | 3.91 | RSN587_NEWZEAL_A-MAT083 |
| 827 | "Cape Mendocino" | 1992 | 7.01 | 13.18 | RSN827_CAPEMEND_FOR000 |
| 3750 | "Cape Mendocino" | 1992 | 7.01 | 8.34 | RSN3750_CAPEMEND_LFS270 |
| 3757 | "Landers" | 1992 | 7.28 | 9.91 | RSN3757_LANDERS_NPF090 |

Table 2 – Ground motion selection and scaling for the rocking model

| Record Sequence Number | Earthquake Name | Year | Magnitude | Scale Factor | Record |
|---|---|---|---|---|---|
| 15 | "Kern County" | 1952 | 7.36 | 9.44 | RSN15_KERN_TAF021 |
| 292 | "Irpinia_ Italy-01" | 1980 | 6.9 | 6.62 | RSN292_ITALY_A-STU000 |
| 313 | "Corinth_ Greece" | 1981 | 6.6 | 6.34 | RSN313_CORINTH_COR--L |
| 570 | "Taiwan SMART1(45)" | 1986 | 7.3 | 9.81 | RSN570_SMART1.45_45C00EW |
| 582 | "Taiwan SMART1(45)" | 1986 | 7.3 | 10.59 | RSN582_SMART1.45_45O08EW |
| 587 | "New Zealand-02" | 1987 | 6.6 | 5.28 | RSN587_NEWZEAL_A-MAT083 |

## 2.1. Shear frame model

First, a three-storey steel shear frame was examined using three two-node-link elements. To model the lateral behaviour of the shear frame, the two-node-link elements are defined to provide lateral stiffness only, while maintaining sufficiently stiff in axial and rotational directions. The storey height is assumed to be 118 inches at all levels. A macro frame model (i.e. one-bay-three-storey steel frame) is constructed which is pinned at the base and the nonlinear behaviour is represented using the lumped plasticity concept with the rotational springs whose properties are calibrated using steel02 material in OpenSees, according to the experimental results in [14] (Fig. 1 and Fig. 2). It is assumed the nonlinear behaviour of the macro frame is concentrated at the hinges and the columns remain elastic. Therefore, the columns are modelled using elastic beam-column elements with the section properties in line with [14], and beams are modelled using sufficiently stiff elastic beam-column elements. The fracture phenomenon is ignored. The shear behaviour of the two-node link elements is defined such that it is identical to the lateral behaviour of the macro frame model that adopts the rotational spring properties calibrated against [14]. It should be noted that the contribution to the hinge rotation by the column deformation should be eliminated from the total hinge rotation. The two-node-link elements also adopt the steel02 material to maintain the consistency with the macro model. In order to maintain the reasonable modal properties of the model, the nodal mass is assumed as $0.3 kip \cdot sec^2/in$ at the first and second levels, and $0.18 kip \cdot sec^2/in$ at the roof level. Modal analysis shows the periods of the shear frame model are 0.550sec, 0.188sec and 0.120sec, respectively for the three modes. 2.5% Rayleigh damping is applied based on the first and third modes of the frame. To examine the performance of ANNs in the presence of only the material nonlinearity, P-delta effect is neglected for the proposed shear frame.

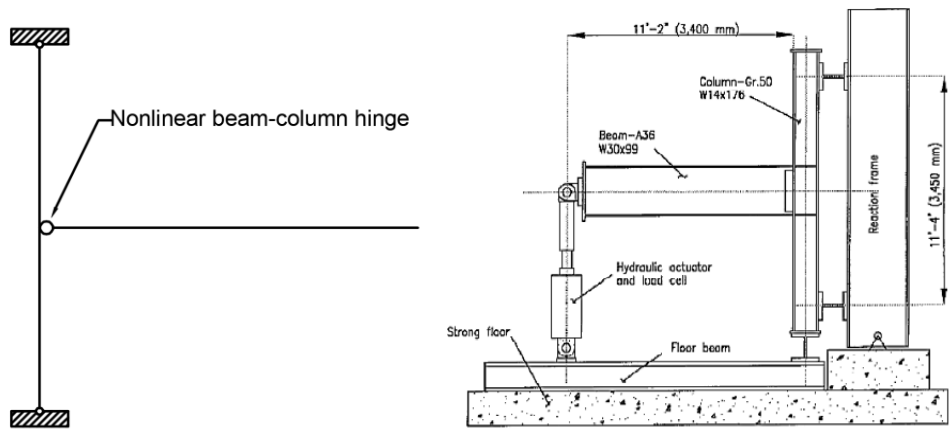

Fig. 1 – Nonlinear hinge calibration and experimental setup [14].

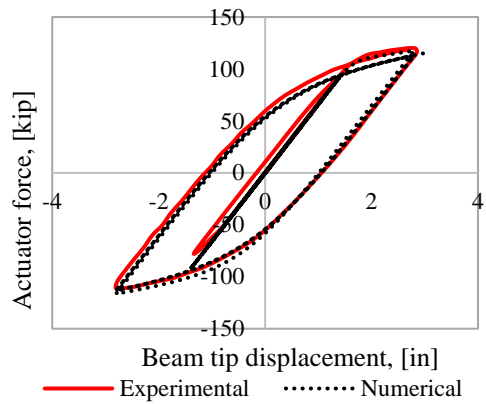

Fig. 2 – Comparison of numerical estimation and experimental results

## 2.2. Solitary rocking model

Second, a model of a rectangular rocking body is constructed (Fig. 3). Modelling of rocking surface and the energy dissipation strategy follow the general methodology defined in [15]. In this study, $2w = 4m$, $2h = 12m$ (Fig. 3). As a result, the geometric angular constant, $\alpha = 0.322\ rads$. The rocking surface is constructed using zero-length section elements defined in OpenSees. The zero-length section elements employ the two-fiber section approach which assumes elastic-no-tension material with sufficiently high compressive stiffness. The fiber stiffness should be assigned such that a further increase in the stiffness does not result in evident changes in the rocking response. The rocking body is modelled using elastic beam-column elements with sufficiently high stiffness. Corotational geometric transformation is selected. In addition, it is assumed the rocking body does not experience sliding during rocking. Therefore, the contribution to the rocking amplitude only comes from the rotation about the pivots. More modelling assumptions can be referred to [15].

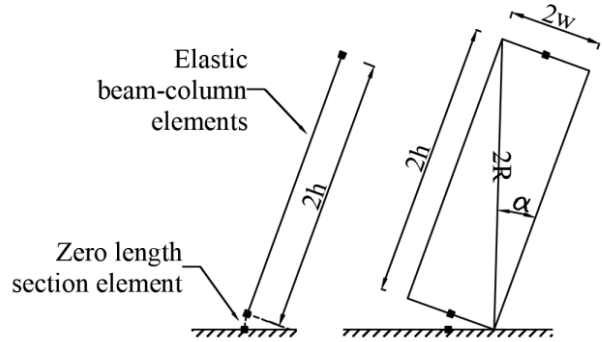

Fig. 3 – Numerical model of the solitary rocking block.

Nodal mass distribution is considered following [15]. The moment of inertia of the proposed rectangular rocking body around one of its pivot points can be computed as $\frac{4mR^2}{3}$. If the translational masses are assigned to the distributed nodes along the elastic beam-column elements of the rocking body, the corresponding moment of inertia can be evaluated as $mR^2\left(1 + \frac{1}{3}\cos^2\alpha\right)$. Therefore, the total supplemental rotational mass determined as $\frac{1mR^2}{3}\sin^2\alpha$, should be evenly distributed over the nodes on the rocking body. Details of the nodal mass distribution can also be referred to [16].

As suggested by [15], neither the classical nor the non-classical damping is applied to the rocking body and the rocking surface. Instead, a numerical energy dissipation method is employed. For these purposes, a common time-stepping algorithm, Hilber–Hughes–Taylor (HHT) [17] is considered. the energy dissipation factor assumes 0.667 in OpenSees HHT algorithm. The time step employed is $10^{-4}$sec because modelling the effects of the high-frequency and small-amplitude shock waves demands a relatively small time step [15].

## 3. Adaptive artificial neural networks

### 3.1. General architecture

In this study, a series of fully connected feedforward neural networks (FCNNs) was implemented with an adaptive training scheme, to predict the displacement response of structures under earthquake shaking. In the substructure of the FCNNs, every node in the subsequent layer is connected with all the nodes in the preceding layer with different weights and biases that are integrated with the desired activation function. This fully connected structure is one of the most fundamental neural networks, which can provide strong capability of nonlinear mapping while preventing significant information loss. It should be noted that unlike the conventional ANNs practice, the proposed ANNs do not have a fixed number of nodes and layers owing to the adaptive training scheme. At the end of the training process, the network with appropriate architecture is automatically established. This is appealing because not only can it prevent over-fitting, but also reduce the computational cost through preventing the ANNs from growing to a large and complex architecture with extra nodes and layers to solve the problems of interest. Details of training are introduced in the subsequent sections. The schematic of ANNs is depicted in Fig. 4, where $p(t)$, $p(t-1)$, $p(t-2)$ represent the earthquake input at the time instant $t$, $t-1$ and $t-2$, respectively. The terms, $O_1(t)$, $O_2(t)$, $O_3(t)$ represent the predicted displacement response of the three floors, respectively, by the ANNs. Similarly, $O_{1,gt}(t)$, $O_{2,gt}(t)$, $O_{3,gt}(t)$ are the ground truth displacement of each floor at the instant $t$, which are generated from the FE models aforementioned. In addition, the Hyperbolic Tangent function, $tanh$ is selected as the activation function.

In order to approximate the dynamic response of the shear frame model which consists of 3 DOFs, $n$ is taken as 3. Consequently, the ANNs take 9 (i.e. $2n+3$ where $n=3$) inputs which include 3 ground motion values, and response parameters at the present and previous steps at 3 floor levels, passing through the hidden layers, and finally produce 3 outputs that contain the floor response parameters at the subsequent time step.

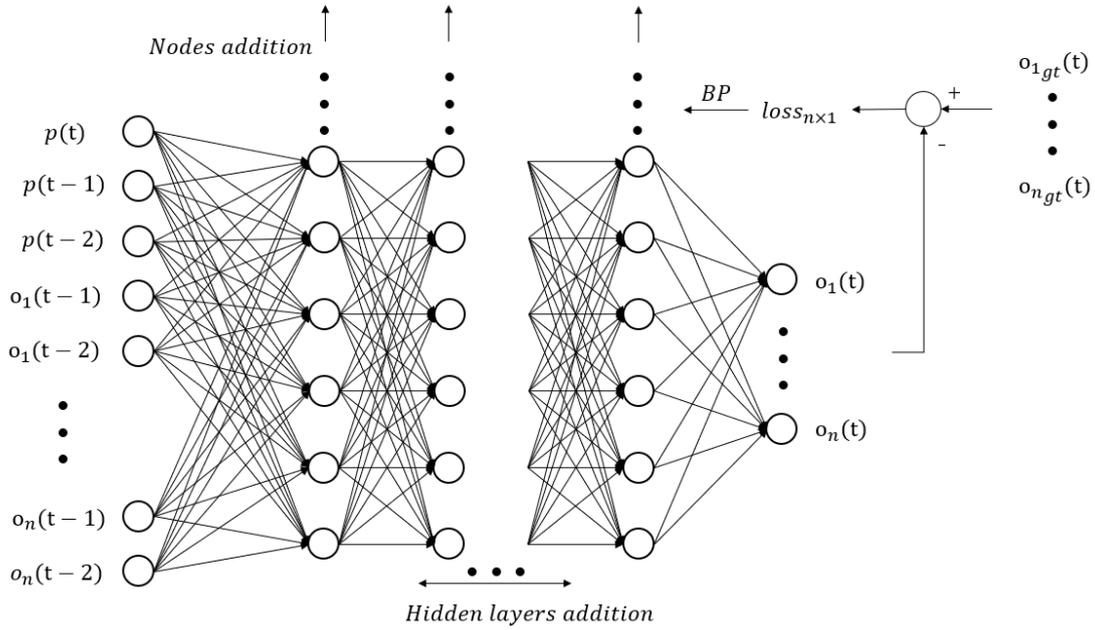

Fig. 4 – The architecture of the adaptive ANNs

### 3.2. Adaptive training scheme

#### 3.2.1. Dynamic learning procedures

The proposed ANNs utilizes two ground motions for training, while the errors are cross-validated to fine-tune the best possible set of weights and biases. The training is achieved with the common methods, Stochastic Gradient Descent (SGD) and Back Propagation (BP). The loss function is defined as the arithmetic difference between the prediction and the ground truth:

$$Loss(t) = Groundtruth(t) - Output(t) \qquad (1)$$

The training time-series data are traversed once in a loop through SGD and BP to determine the adjusted weights and biases. This process is repeated many times for a single training case (i.e. the response under one single ground motion). The objective of the training is to minimize the prediction error of the whole time series. Supposing there are $n$ time instants in a single training series, the error of the entire time series can be calculated by:

$$Error(epoch) = \sum_{t=1}^{n} Loss(t) = \sum_{t=1}^{n} [Groundtruth(t) - Output(t)] \qquad (2)$$

The learning rate determines the speed at which the networks approach the optimal point. To reduce the workload of supervised manual hyperparameter tuning, the learning rate is dynamically adjusted during training. The learning process starts from a relatively large learning rate, 0.5. As the training goes on, the networks gradually approach the minimal where the error may oscillate around, and the learning rate will be halved to facilitate training. The training is terminated until the loss is reduced to a certain threshold, which is taken as 3% average error for a training series in the present study. This training process enhances the training efficiency and rationally minimizes the distance to the true optimal point.

### 3.2.2. Dynamic generation of nodes and layers

Traditional ANNs are constructed with fixed architecture with a constant number of nodes and hidden layers, and consequently are more likely to get stuck in local minima during training. In this case, the prediction error cannot further decline, even with an extremely small learning rate. It implies that the existing structure of the ANNs has reached its capacity and extra nodes or layers should be added to achieve successful training. This study incorporates two dynamic modifications simultaneously (Fig. 5), including a) widening: addition of one node to every layer of the network, b) deepening: addition of a new layer between the last hidden layer and the output layer, with the equal number of nodes to the existing hidden layers. In comparison to deepening, widening is a more conservative process, because adding a new branch only slightly influences the output, especially when the network already has a relatively large number of nodes. On the other hand, the effect of deepening can be sometimes troublesome, leading to exploding gradients problem. When the ANNs reach the saturation with the existing architecture, the networks will be firstly widened. After multiple iterations of widening operations, if the loss is still above the threshold, the network will be deepened once.

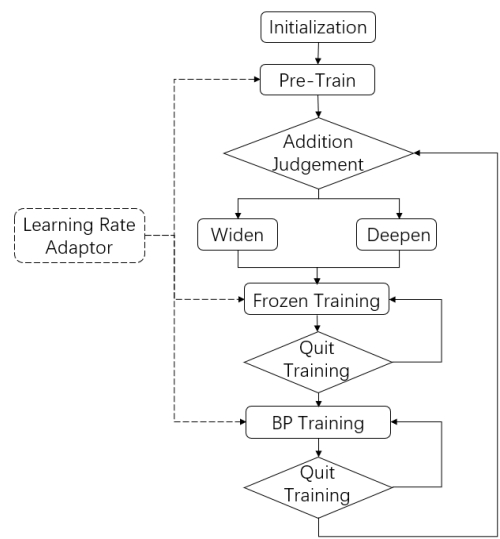

Fig. 5 – Flowchart of adaptive training

The ANNs are initialized with a simple architecture consisting of only 2 hidden layers that have 5 nodes per layer. The ANNs are pretrained prior to enter adaptive training. The optimization loss during the training of ANNs is continuously monitored. If the loss cannot satisfy the desired threshold, the networks enter adaptive training. In the adaptive training, the network is widened by one node at one time. A layer addition is conducted once every 5 widen operations. The nodes and layers are dynamically generated until the loss decreases below the threshold. There are two training modes including a) frozen training mode: the weights and biases of the existing nodes are frozen, while the newly added nodes (from addition of nodes or layers) are trained, and b) normal training mode: all the weights are trained simultaneously. After widening and deepening, the networks should first conduct frozen training before entering the normal training mode. Such training strategy is based on the fact that the random initial weights and biases of the newly added nodes might significantly deviate the ANNs from the optimal state, and therefore require preliminary regularization. In this study, the ANNs are implemented with 10,000 times of frozen training, before proceeding with normal training (Fig. 6). The black nodes represent the existing nodes and the white ones are newly added. The dashed line connections refer to the frozen weights while the solid line connections represent the weights being adjusted through training.

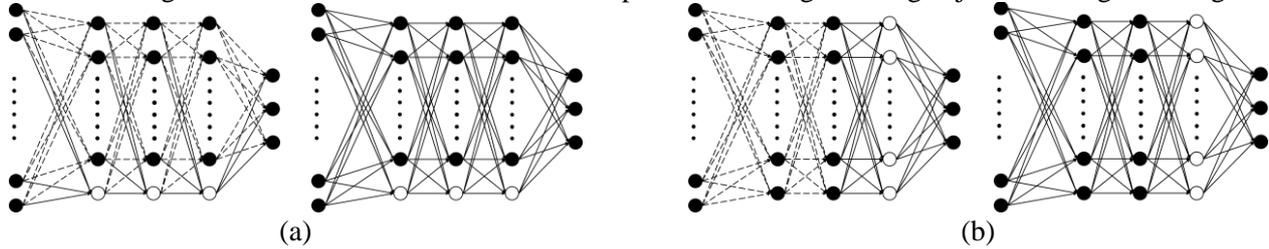

(a)                                                                                              (b)

Fig. 6 – Frozen training mode and normal training mode during (a) dynamic nodes generation (b) dynamic layers generation

## 4. Results

### 4.1. Shear frame model response

Fig. 7 provides the comparison of the sample training results against the finite element responses of the shear frame. The displacement response is defined as the absolute displacement relative to the ground and normalized by the storey height. Evidently, the results of the proposed ANNs are in line with the results of the traditional FE models. In addition, the storey component behaviour is shown in Fig. 8, which indicates the structure has been excited to a highly nonlinear state. The ANNs are trained using 2 ground motions and tested by 9 different ground motions. Table 3 summarizes the training and testing results, which indicates the proposed adaptive ANNs yield great accuracy in predicting the time history response of the nonlinear shear frame structures.

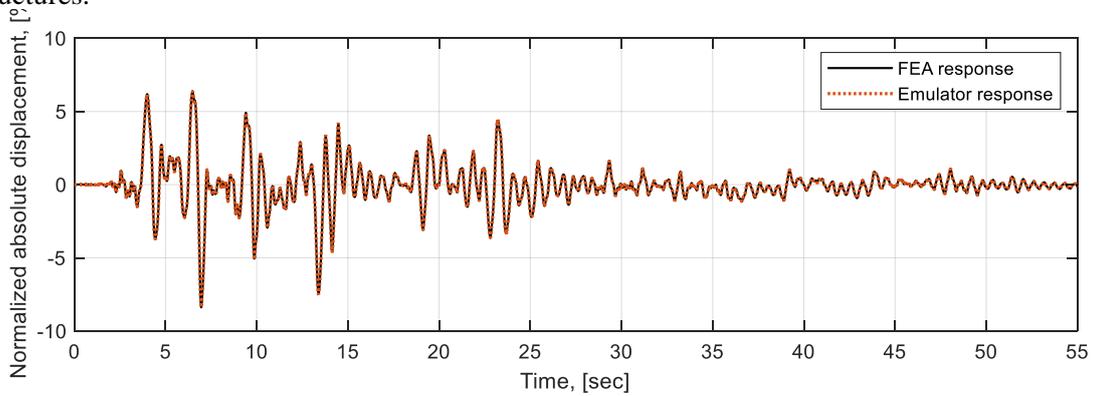

(a)

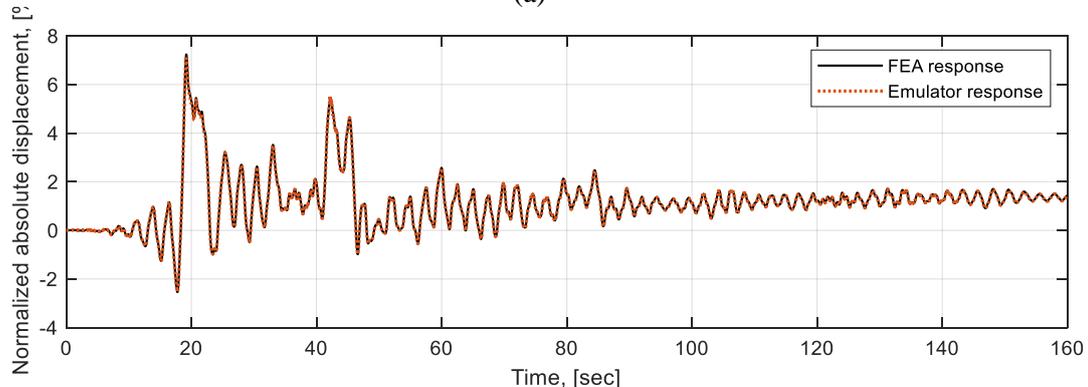

(b)

Fig. 7 – Comparison of the 1$^{st}$-floor response between the trained adaptive ANNs and FE models under (a) RSN15, and (b)RSN 292

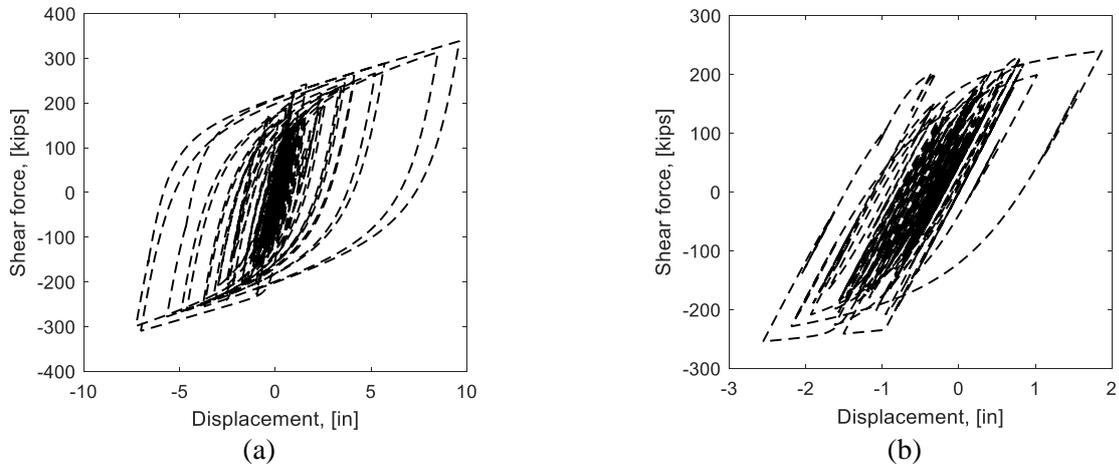

Fig. 8 – Relationship between the shear force and the displacement of the component under RSN15 ground motion at (a) 1st floor, and (b) 2nd floor

Table 3. Average error rate at each floor

| Ground motions | 1st floor | 2nd floor | 3rd floor | |
|---|---|---|---|---|
| RSN15 | 1.8% | 1.5% | 0.8% | Training |
| RSN292 | 1.3% | 0.9% | 1.1% | |
| RSN164 | 2.3% | 1.8% | 0.8% | Testing |
| RSN143 | 4.2% | 3.6% | 3.4% | |
| RSN313 | 2.0% | 1.9% | 0.8% | |
| RSN570 | 1.6% | 1.9% | 0.9% | |
| RSN582 | 1.9% | 1.4% | 0.8% | |
| RSN587 | 4.9% | 5.5% | 3.5% | |
| RSN827 | 3.4% | 3.2% | 3.2% | |
| RSN3750 | 1.6% | 1.0% | 1.0% | |
| RSN3757 | 1.4% | 0.9% | 1.0% | |

## 4.2. Solitary rocking model response

Similarly, the capability of the proposed ANNs in predicting the rocking response is validated in this section. Fig. 9 shows the comparison of the ANNs prediction and the FE results during training. The response is represented by the rocking amplitude normalized by the geometric angular constant, $\alpha$. It is evident that the ANNs are successfully trained to predict the rocking response under these two ground motions. New ground motions are fed into the trained ANNs and the sample testing results are provided in Fig. 10. In general, the testing results substantiate the robustness of the trained ANNs, and the peak response is well captured in the cases of all the ground motions. The average error under each excitation is below 5% and are summarized in Table 4.

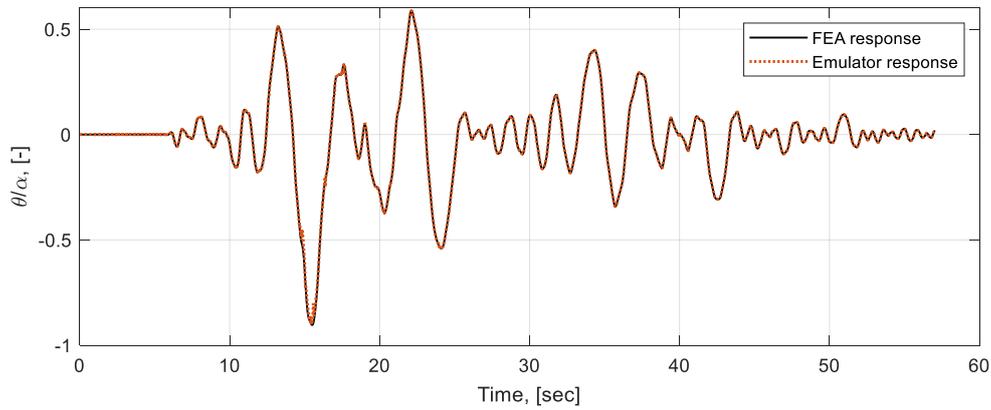

(a)

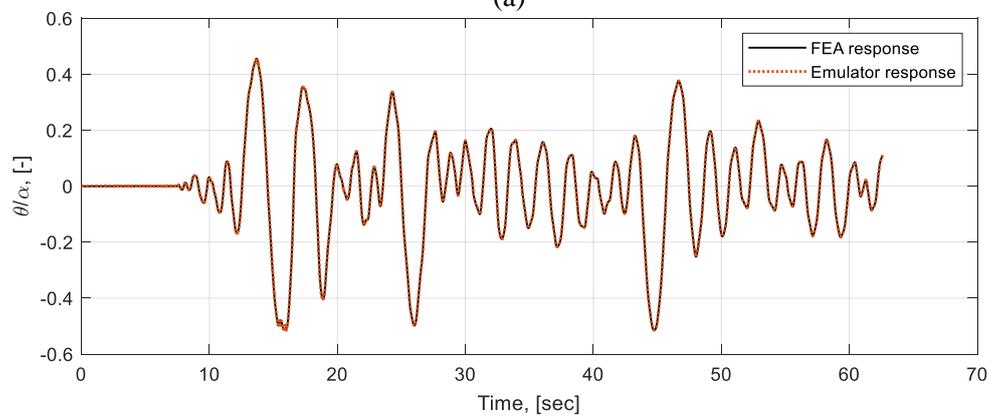

(b)

Fig. 9 – Training results: normalized rotation time-history response under (a) RSN 570, and (b) RSN582

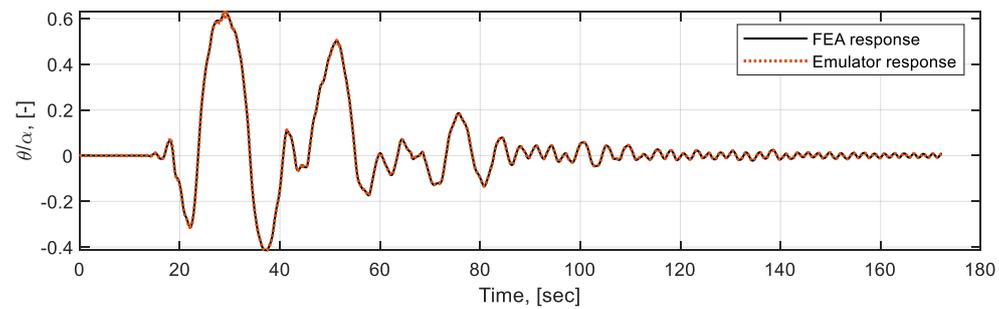

(a)

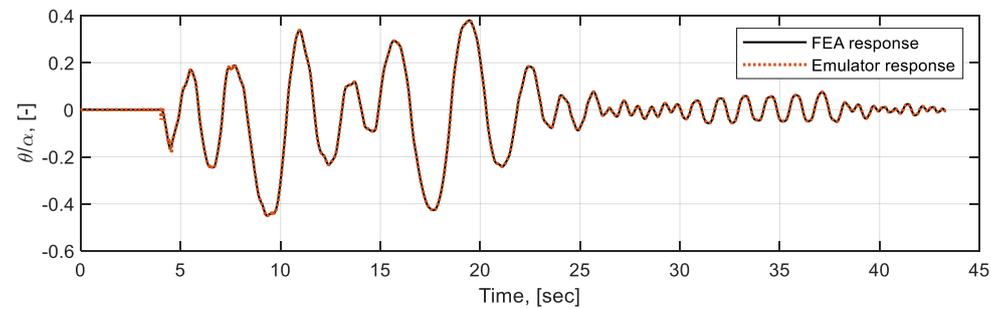

(b)

Fig. 10 – Sample testing results: normalized rotation time-history response under (a) RSN 292, and (b) RSN 313

Table 4. Average error rate

| Ground motions | Normalized rocking amplitude, $\theta/\alpha$ | |
|---|---|---|
| RSN570 | 0.97% | Training |
| RSN582 | 0.41% | |
| RSN15 | 1.97% | Testing |
| RSN292 | 0.53% | |
| RSN313 | 0.48% | |
| RSN587 | 4.70% | |

In addition, to examine the computational efficiency of the proposed ANNs against the FE models, 100 time-history analysis are performed by the rocking models in OpenSees, and the proposed ANNs in MATLAB 2019a. The GPUs acceleration is enabled during the testing phase in MATLAB with the aid of neural network toolbox and parallel computing toolbox. The total time is recorded. The training is implemented in MATLAB R2019a by Alienware Aurora R8 (a Core i7-9700K@ 3.60 GHz, 16 GB DDR4 memory and 8 GB memory GeForce RTX 2070 GPU). The results show that the trained ANNs can reduce the computational time by 43% in comparison to the FE models, when the GPUs acceleration is activated. This indicates the development of rocking spectrum can be achieved much more efficiently where a large number of input motions with various frequency parameters are required to produce the spectrum.

## 5. Conclusion

Nonlinear time-history analysis of structures under earthquake demands high computational power. Recent advances in computer graphical power provide a more efficient solution in approximating structural response using artificial neural networks. In this study, the adaptive training schemes are implemented for the ANNs and the performance of ANNs is validated against the traditional finite element modelling which includes both geometric and material nonlinearities. Overall, the response evaluation by the proposed ANNs is in good agreement with the finite element approaches. In the second case study, the trained ANNs yield higher computational efficiency using the parallel computing technique enabled by the recently-developed GPUs. Although there exist minor discrepancies in a few local regions, the peak response of rocking is well captured. Therefore, the proposed ANNs can be extended to facilitate the determination of the response spectra of rigid rocking bodies, where the structures need to be examined under a large number of input motions with various frequency parameters.

## 6. Acknowledgements

The authors would like to acknowledge the funding provided by the International Joint Research Laboratory of Earthquake Engineering (ILEE), National Natural Science Foundation of China (grant number: 51778486), Natural Sciences and Engineering Research Council (NSERC), China Scholarship Council. Any opinions, findings, and conclusions or recommendations expressed in this paper are those of the authors.